\documentclass[11pt,a4paper]{article}
\usepackage[hyperref]{acl2021}
\usepackage{times}
\usepackage{latexsym}

\usepackage{xcolor}
\usepackage{graphicx}

\usepackage{tabularx}
\usepackage{multirow}
\usepackage{booktabs}
\usepackage{amsmath}
\usepackage{amssymb}
\usepackage{float}

\usepackage{microtype}

\aclfinalcopy 

\title{Automatic Fake News Detection: Are Models Learning to Reason?}

\author{Casper Hansen \\
  University of Copenhagen \\
  \texttt{c.hansen@di.ku.dk} \\\And
  Christian Hansen \\
  University of Copenhagen \\
  \texttt{chrh@di.ku.dk} \\\And
  Lucas Chaves Lima \\
  University of Copenhagen \\
  \texttt{lcl@di.ku.dk} \\}
\date{}

\begin{document}
\maketitle
\begin{abstract}
Most fact checking models for automatic fake news detection are based on reasoning: given a claim with associated evidence, the models aim to estimate the claim veracity based on the supporting or refuting content within the evidence. When these models perform well, it is generally assumed to be due to the models having learned to reason over the evidence with regards to the claim. In this paper, we investigate this assumption of reasoning, by exploring the relationship and importance of both claim and evidence. Surprisingly, we find on political fact checking datasets that most often the highest effectiveness is obtained by utilizing only the evidence, as the impact of including the claim is either negligible or harmful to the effectiveness. This highlights an important problem in what constitutes evidence in existing approaches for automatic fake news detection.
\end{abstract}

\section{Introduction}
Misinformation is spreading at increasing rates \cite{vosoughi2018spread}, particularly online, and is considered a highly pressing issue by the World Economic Forum \cite{howell2013}. To combat this problem, automatic fact checking, especially for estimating the veracity of potential fake news, have been extensively researched \cite{hassan2017toward,hansen2019neural,thorne2018automated,10.1007/978-3-030-15719-7_41,allein2020time,popat2018declare,augenstein2019multifc}. Given a claim, most fact checking systems are \emph{evidence-based}, meaning they utilize external knowledge to determine the claim veracity. Such external knowledge may consist of previously fact checked claims \cite{shaar2020known}, but it typically consists of using the claim to query the web through a search API to retrieve relevant hits. 
While including the evidence in the model increases the effectiveness over using only the claim, existing work has not focused on the predictive power of isolated evidence, and hence whether it assists the model in enabling better reasoning.

In this work we investigate if fact checking models learn reasoning, i.e., provided a claim and associated evidence, whether the model determines the claim veracity by reasoning over the evidence. If the model learns reasoning, we would expect the following proposition to hold: \textit{A model using both the claim and evidence should perform better on the task of fact checking compared to a model using only the claim or evidence}. 
If a model is only given the claim as input, it does not necessarily have the information needed to determine the veracity. Similarly, if the model is only given the evidence, the predictive signal must come from dataset bias or the differences in the evidence obtained from claims with varying veracity, as it otherwise corresponds to being able to provide an answer to an unknown question. In our experimental evaluation on two political fact checking datasets, across multiple types of claim and evidence representations, we find the evidence provides a very strong predictive signal independent of the claim, and that the best performance is most often obtained while entirely ignoring the claim. This highlights that fact checking models may not be learning to reason, but instead exploit an inherent signal in the evidence itself, which can be used to determine factuality independent of using the claim as part of the model input. This highlights an important problem in what constitutes evidence in existing approaches for automatic fake news detection. We make our code publicly available at \url{https://github.com/casperhansen/fake-news-reasoning}.


\section{Related Work}
Automatic fact checking models include deep learning approaches, based on contextual and non-contextual embeddings, which encode the claim and evidence using RNNs or Transformers \cite{shaar2020known,10.1007/978-3-030-15719-7_41,allein2020time,popat2018declare,augenstein2019multifc,hassan2017toward}, and non-deep learning approaches \cite{P17-2067,C18-1287}, which uses hand-crafted features or bag-of-word representations as input to traditional machine learning classifiers such as random forests, SVM, and MLP \cite{mihalcea2009lie,C18-1287,DBLP:conf/naacl/BalyMGMMN18,DBLP:journals/corr/abs-1809-00509}. 

Generally, models may learn to memorize artifact and biases rather than truly learning~\cite{gururangan-etal-2018-annotation,moosavi-strube-2017-lexical,agrawal-etal-2016-analyzing}, e.g., from political individuals often leaning towards one side of the truth spectrum. Additionally, language models have been shown to implicitly store world knowledge \cite{roberts2020much}, which in principle could enhance the aforementioned biases. To this end, we design our experimental setup to include representative fact checking models of varying complexity (from simple term-frequency based representations to contextual embeddings), while always evaluating each trained model on multiple different datasets to determine generalizability.


\section{Methods}
\textbf{Problem definition.} In automatic fact checking of fake news we are provided with a dataset of $ D = \{(c_1, e_1, y_1), ..., (c_n, e_n, y_n)\}$, where $c_i$ corresponds to a textual claim, $e_i$ is evidence used to support or refute the claim, and $y_i$ is the associated truth label to be predicted based on the claim and evidence. Following current work on fact checking of fake news \cite{hassan2017toward,thorne2018automated,10.1007/978-3-030-15719-7_41,allein2020time,popat2018declare,augenstein2019multifc}, we consider the evidence to be a list of top-10 search snippets as returned by Google search API when using the claim as the query. Note that while additional metadata may be available--such as speaker, checker, and tags--this work focuses specifically on whether models learn to reason based on the combination of claim and evidence, hence we keep the input representation to consist only of the latter.

\textbf{Overview.} In the following we describe the different models used for the experimental comparison (Section~\ref{s:exp-eval}), which consists of models based on term frequency (term-frequency weighted bag-of-words \cite{salton1988term}), word embeddings (GloVe word embeddings \cite{pennington2014glove}), and contextual word embeddings (BERT \cite{devlin-etal-2019-bert}). These representations are chosen as to include the typical representations most broadly used among past and current NLP models.

\begin{table*}[]
    \centering
    \resizebox{0.92\linewidth}{!}
     {
\begin{tabular}{l|cc|cc||cc|cc}
\toprule
    &  \multicolumn{4}{c||}{Train: Snopes} & \multicolumn{4}{c}{Train: PolitiFact} \\
    & \multicolumn{2}{c|}{Within dataset} & \multicolumn{2}{c||}{Out-of dataset} & \multicolumn{2}{c|}{Within dataset} & \multicolumn{2}{c}{Out-of dataset} \\
    & \multicolumn{2}{c|}{Eval: Snopes} & \multicolumn{2}{c||}{Eval: PolitiFact} & \multicolumn{2}{c|}{Eval: PolitiFact} & \multicolumn{2}{c}{Eval: Snopes} \\
    RF ($\sim$13 seconds) & $\textrm{F1}_{\textrm{micro}}$ & $\textrm{F1}_{\textrm{macro}}$ & $\textrm{F1}_{\textrm{micro}}$ & $\textrm{F1}_{\textrm{macro}}$ & $\textrm{F1}_{\textrm{micro}}$ & $\textrm{F1}_{\textrm{macro}}$ & $\textrm{F1}_{\textrm{micro}}$ & $\textrm{F1}_{\textrm{macro}}$ \\ \midrule
Claim & 0.473 & 0.231 & \underline{\textbf{0.273}} & \underline{0.223} & 0.254 & 0.255 & 0.546 & \underline{0.243} \\
Evidence & 0.504 & \underline{0.280} & 0.244 & 0.195 & 0.301 & 0.299 & \underline{\textbf{0.597}} & 0.232 \\
Claim+Evidence & \underline{0.550} & 0.271 & 0.245 & 0.190 & \underline{0.310} & \underline{0.304} & 0.579 & 0.207 \\
\midrule
\multicolumn{9}{l}{LSTM ($\sim$12 minutes, 888K parameters)} \\
\midrule
Claim & 0.408 & 0.243 & 0.260 & \underline{\textbf{0.228}} & 0.237 & 0.237 & \underline{0.565} & 0.221 \\
Evidence & 0.495 & \underline{0.253} & \underline{0.262} & 0.208 & \underline{0.290} & \underline{0.295} & 0.550 & \underline{0.273} \\
Claim+Evidence & \underline{0.529} & \underline{0.253} & 0.258 & 0.189 & 0.288 & 0.294 & 0.509 & 0.256 \\
\midrule
\multicolumn{9}{l}{BERT ($\sim$264 minutes, 109M parameters)} \\
\midrule
Claim & 0.533 & 0.312 & \underline{0.249} & 0.209 & 0.275 & 0.282 & 0.550 & 0.273 \\
Evidence & 0.531 & \underline{\textbf{0.321}} & \underline{0.249} & \underline{0.224} & \underline{\textbf{0.351}} & \underline{\textbf{0.359}} & \underline{0.577} & \underline{\textbf{0.286}} \\
Claim+Evidence & \underline{\textbf{0.556}} & 0.313 & 0.231 & 0.191 & 0.285 & 0.292 & 0.564 & 0.259 \\
\bottomrule
\end{tabular}}
    \caption{Evaluation using micro and macro F1. Per column, the best score per method is underlined and the best score across all methods is highlighted in bold. We report the training time and number of model parameters, for Claim+Evidence on PolitiFact, in the parentheses. RF is trained on 5 cores and neural models on a Titan RTX.}
    \label{tab:results}
\end{table*}

\textbf{Term-frequency based Random Forest}.
We construct a term-frequency weighted bag-of-words representation per sample based on concatenating the text content of the claim and associated evidence snippets. We train a Random Forest~\cite{breiman2001random} as the classifier using the Gini impurity measure. In the setting of only using either the claim or evidence snippets as the input, only the relevant part is used for constructing the bag-of-words representation.

\textbf{GloVe-based LSTM model}.
We adapt the model by \citet{augenstein2019multifc}, which originally was proposed for multi-domain veracity prediction. Using a pretrained GloVe embedding \cite{pennington2014glove}\footnote{\url{http://nlp.stanford.edu/data/glove.840B.300d.zip}}, claim and snippet tokens are embedded into a joint space. We encode the claim and snippets using an attention-weighted bidirectional LSTM~\cite{hochreiter1997long}:
\begin{align} \label{eq:lstm-claim-encode}
    h_{c_i} &= \textrm{attn}\left( \textrm{BiLSTM}(c_i) \right) \\
    h_{e_{i,j}} &= \textrm{attn}\left( \textrm{BiLSTM}(e_{i,j}) \right)
\end{align}
where $\textrm{attn}(\cdot)$ is a function that learns an attention score per element, which is normalized using a softmax, and returns a weighted sum. We combine the claim and snippet encodings using using the matching model by \citet{mou2016natural} as:
\begin{align}
    s_{i,j} = \left[ h_{c_i} \;;\; h_{e_{i,j}} \;;\; h_{c_i} - h_{e_{i,j}} \;;\; h_{c_i} \cdot h_{e_{i,j}} \right]
\end{align}
where ''$;$'' denotes concatenation. The joint claim-evidence encodings are attention weighted and summed, projected through a fully connected layer into $\mathbb{R}^L$, where $L$ is the number of possible labels:
\begin{align} 
    o_i &= \textrm{attn}([s_{i,1} \;;\; ... \;;\; s_{i,10}]) \\
    p_i &= \textrm{softmax} \left( \textrm{FC}( o_i ) \right) \label{eq:lstm-final}
\end{align}
Lastly, the model is trained using cross entropy as the loss function. In the setting of using only the claim as the input (i.e., without the evidence), then $h_{c_i}$ is used in Eq.~\ref{eq:lstm-final} instead of $o_i$. If only the evidence is used, then an attention weighted sum of the evidence snippet encodings is used in Eq.~\ref{eq:lstm-final} instead of $o_i$.

\textbf{BERT-based model}.
In a similar fashion to the LSTM model, we construct a model based on BERT~\cite{devlin-etal-2019-bert}\footnote{We use \texttt{bert-base-uncased} from \url{https://huggingface.co/bert-base-uncased}.}, where the \texttt{[CLS]} token encoding is used for claim and evidence representations. Specifically, the claim and evidence snippets are encoded as:
\begin{align}
    h_{c_i} &= \textrm{BERT}(c_i),\; h_{e_{i,j}} = \textrm{BERT}(c_i, e_{i,j}) \\
    h_{e_i} &= \textrm{attn}([h_{e_{i,1}} \;;\; ... \;;\; h_{e_{i,10}}])
\end{align}
where the claim acts as the question when encoding the evidence snippets. Similarly to Eq. \ref{eq:lstm-final}, the prediction is obtained by concatenating the claim and evidence representations and project it through a fully connected layer into $\mathbb{R}^L$:
\begin{align} \label{eq:bert-final}
    p_i = \textrm{softmax}(FC([h_{c_i} \;;\; h_{e_i}]))
\end{align}
where cross entropy is used as the loss function for training the model. In the setting that only the claim is used as input, then only $h_{c_i}$ is used in Eq.~\ref{eq:bert-final}. If only the evidence is used, then $h_{e_{i,j}}$ is computed without including $c_i$, and only $h_{e_i}$ is used in Eq.~\ref{eq:bert-final}.

\section{Experimental Evaluation}\label{s:exp-eval}
\subsection{Datasets}
We focus on the domain of political fact checking, where we use claims and associated evidence from PolitiFact and Snopes, which we extract from the MultiFC dataset \cite{augenstein2019multifc}. Using the claim as a query, the evidence is crawled from Google search API as the search snippets of the top-10 results, and is filtered such that the website origin of a given claim does not appear as evidence. To facilitate better comparison between the datasets, we filter claims with non-veracity related labels\footnote{For PolitiFact we exclude [full flop, half flip, no flip] and for Snopes we exclude [unproven, miscaptioned, legend, outdated, misattributed, scam, correct attribution].}.
The dataset statistics are shown in Table~\ref{tab:statistics}.

\begin{table}[]
    \centering
    \resizebox{1.0\linewidth}{!}
     {
    \begin{tabular}{l|c|p{0.8\linewidth}}
    \toprule
         &  \#Claims & Labels \\ 
    \midrule
        PolitiFact & 13,581 & \small pants on fire! (10.6\%), false (19.2\%), mostly false (17.0\%), half-true (19.8\%), mostly true (18.8\%), true (14.8\%)  \\ \hline
        Snopes & 5,069 & \small false (64.3\%), mostly false (7.5\%), mixture (12.3\%), mostly true (2.8\%), true (13.0\%) \\
    \bottomrule
    \end{tabular}}
    \caption{Dataset statistics.
    }
    \label{tab:statistics}
\end{table}
\begin{figure*}
    \centering
    \begin{minipage}[]{0.95\linewidth}
    \includegraphics[width=0.325\linewidth]{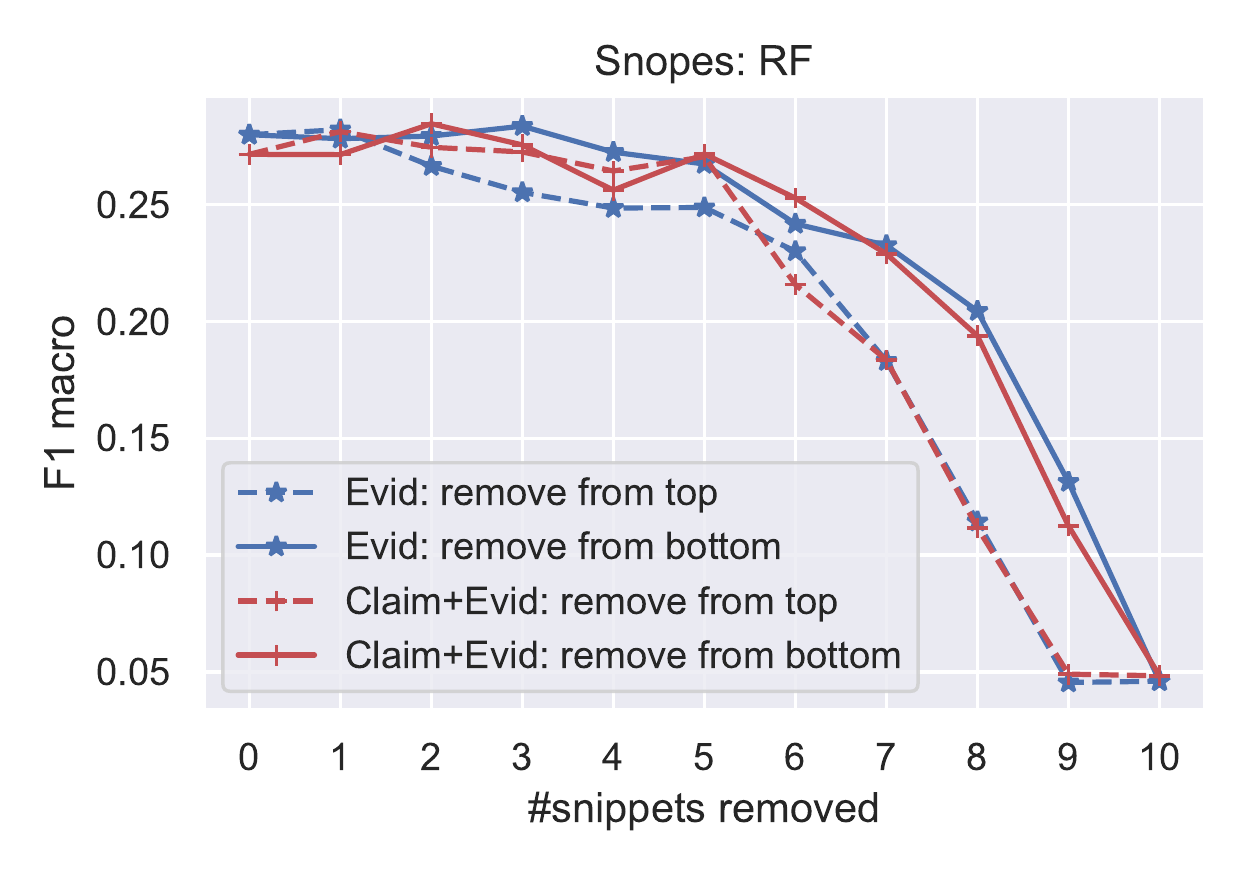}
    \includegraphics[width=0.325\linewidth]{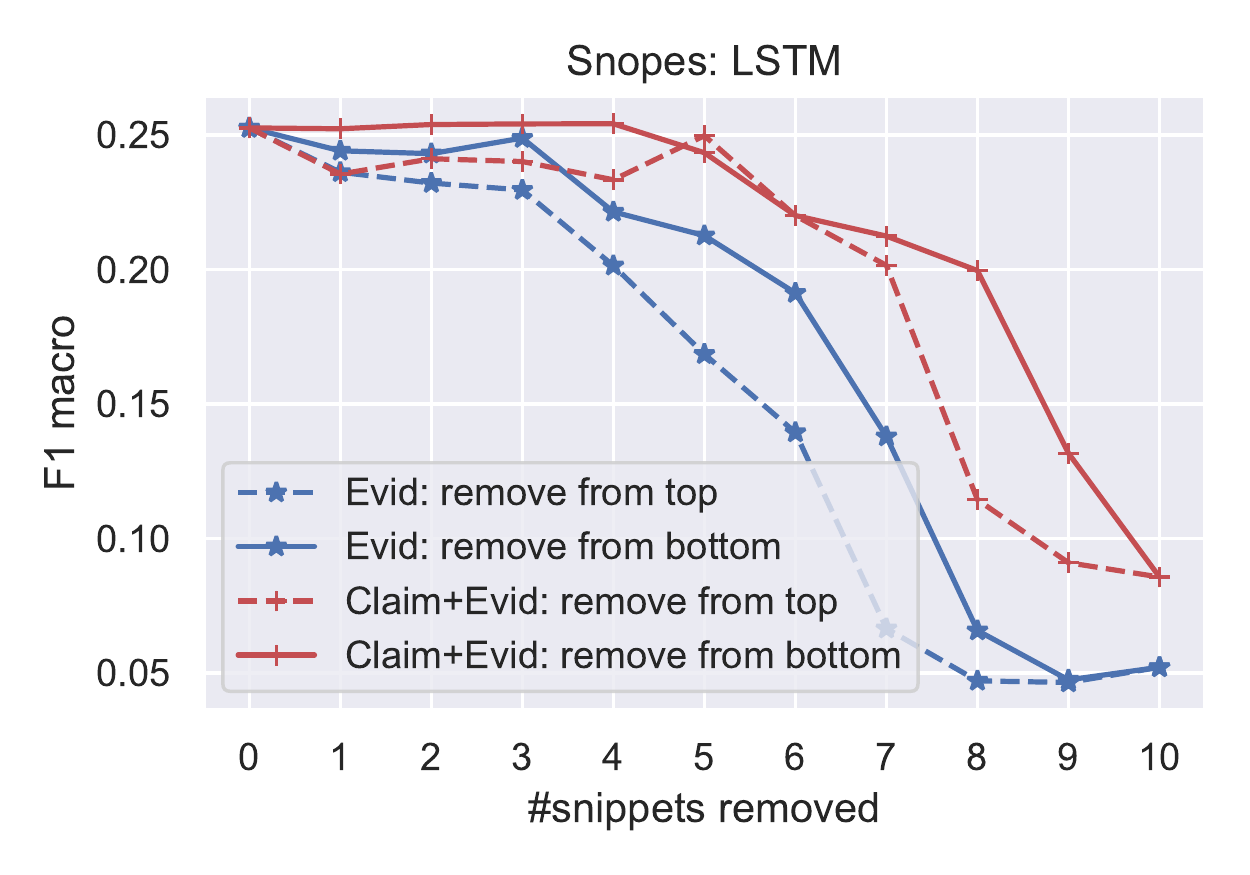}
    \includegraphics[width=0.325\linewidth]{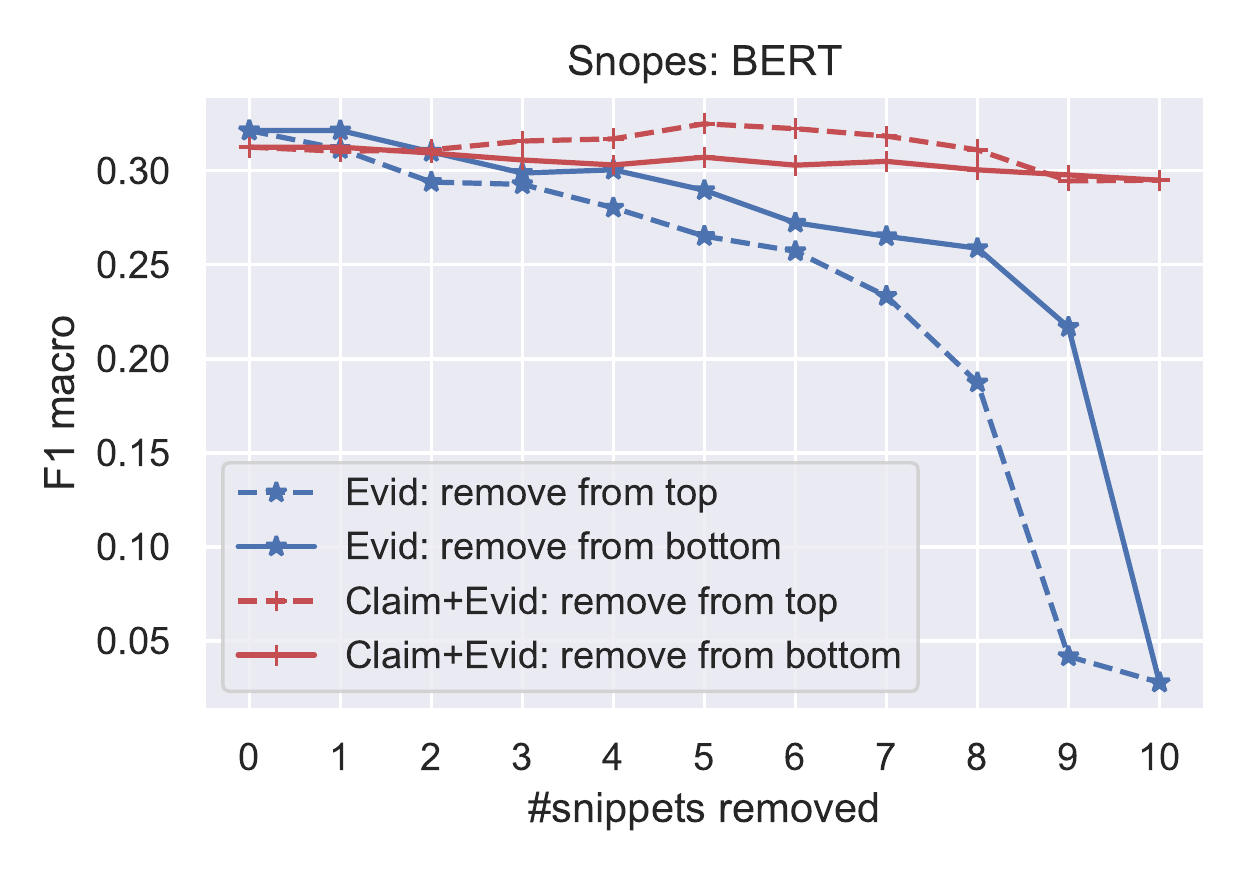}
    \end{minipage}
    \begin{minipage}[]{0.95\linewidth}
    \includegraphics[width=0.325\linewidth]{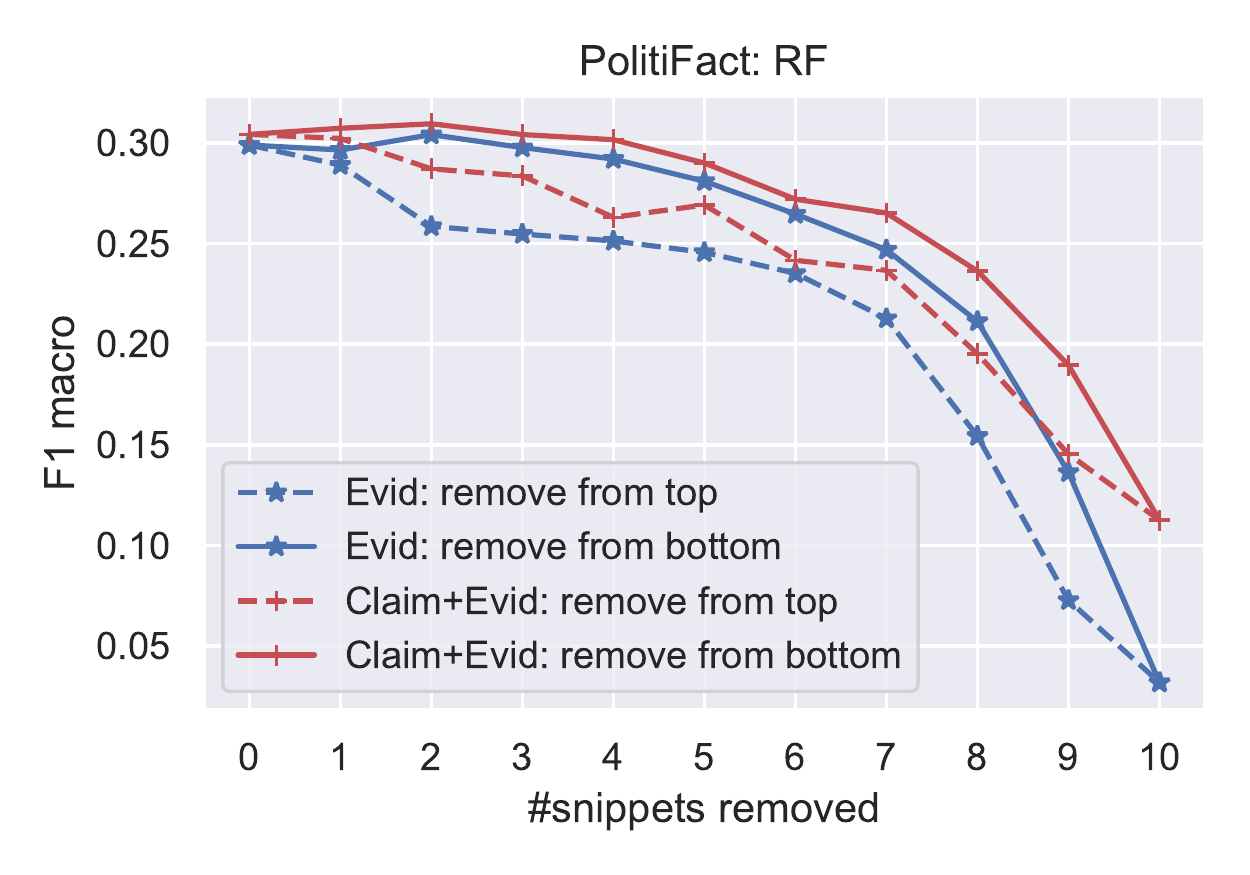}
    \includegraphics[width=0.325\linewidth]{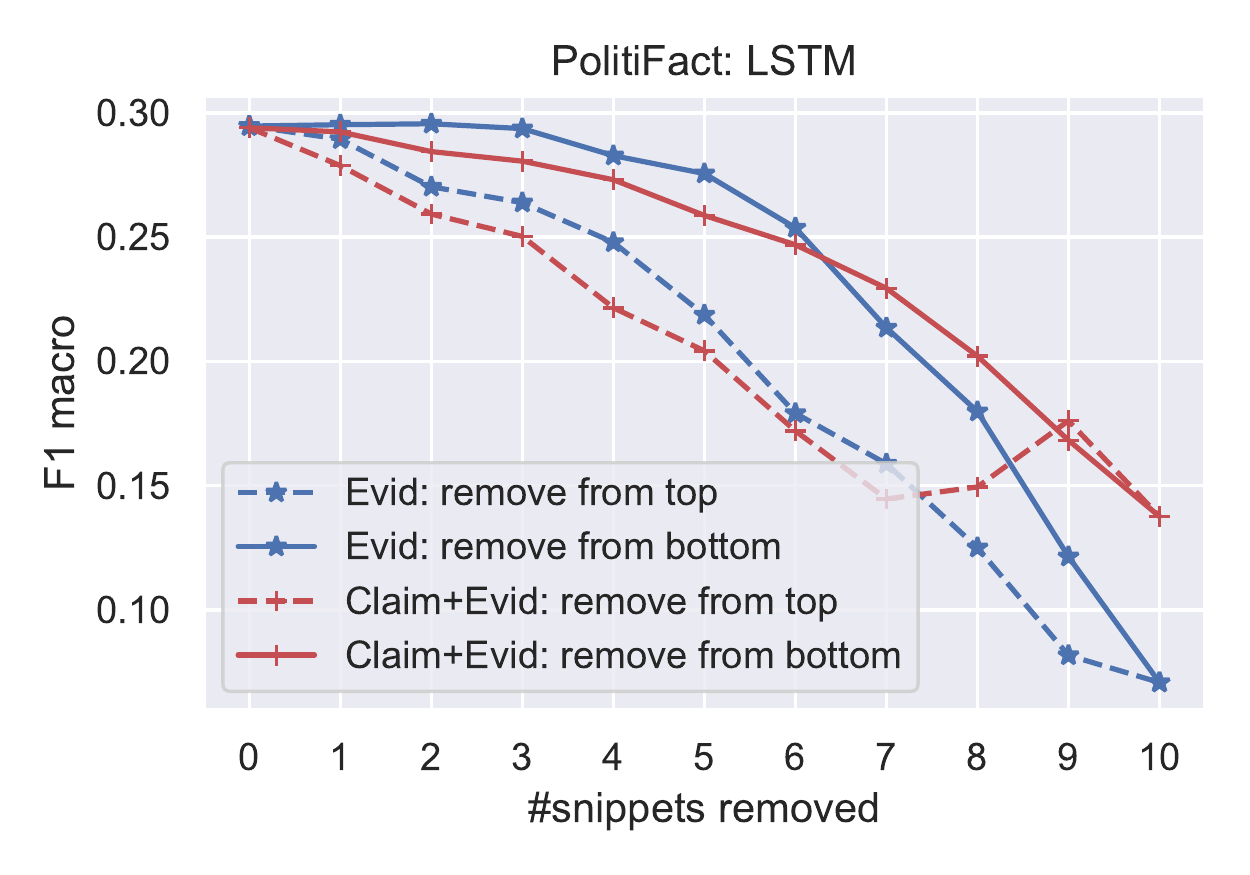}
    \includegraphics[width=0.325\linewidth]{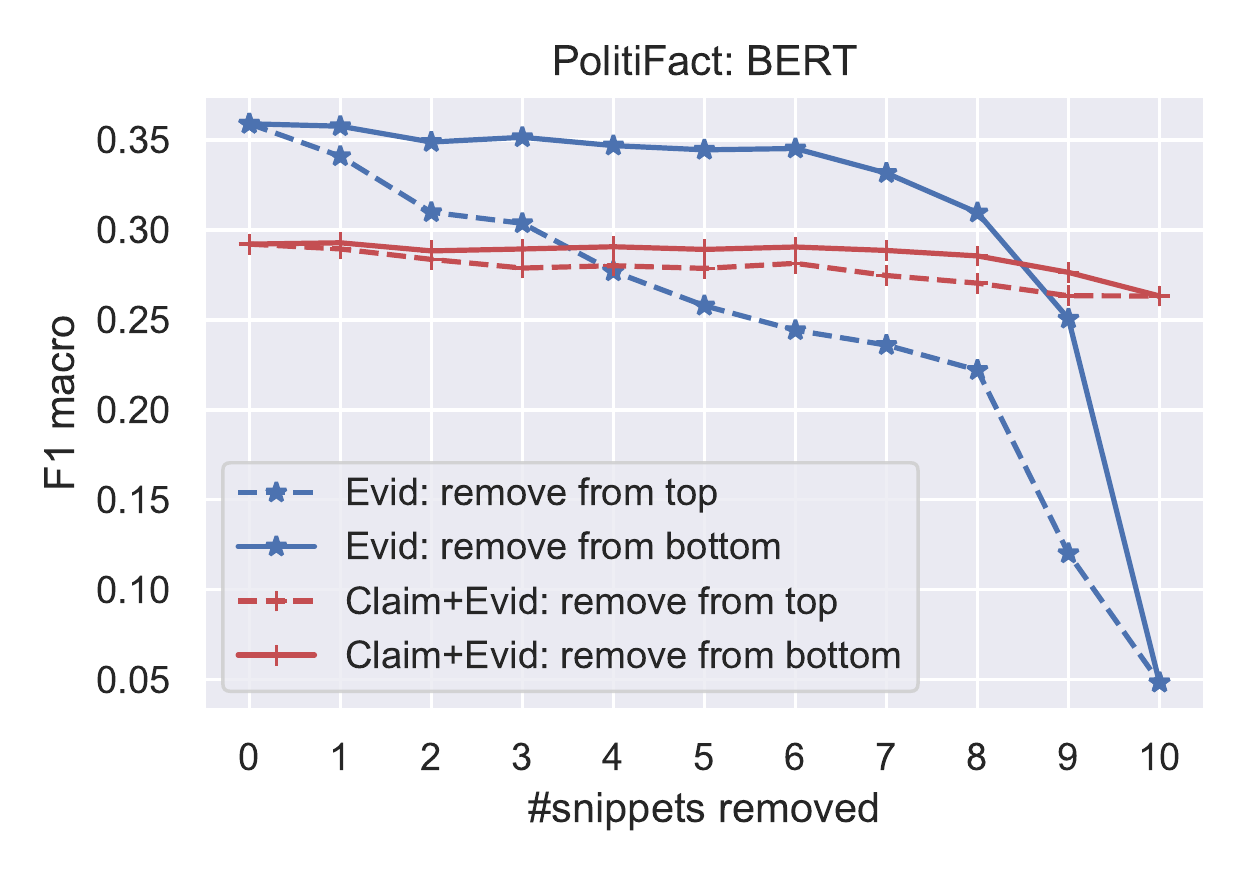}
    \end{minipage}
    \caption{Macro F1 scores when removing evidence from either the top or bottom of the evidence snippet ranking.}
    \label{fig:plots}
\end{figure*}
\subsection{Experimental setup}
Both datasets are split into train/val/test sets using label-stratified sampling (70/10/20\% splits). We tune all models on the validation split, and use early stopping with a patience of 10 for neural models. Following \citet{augenstein2019multifc}, we use micro and macro F1 for evaluation. The models are evaluated on both the within dataset test sets, but also out-of dataset test sets (e.g., a model trained on Snopes is evaluated on both Snopes and PolitiFact). In the out-of dataset evaluation we need the labels to be comparable, hence in that setting we merge ''pants on fire!'' and ''false'' for PolitiFact. 

\section{Tuning details}
In the following, the best overall parameter configurations are underlined. The best configuration is chosen based on the average of the micro and macro F1\footnote{\url{https://scikit-learn.org/stable/modules/generated/sklearn.metrics.f1_score.html}}. For RF, we tune the number of trees from [100,500,\underline{1000}], the minimum number of samples in a leaf from [1,\underline{3},5,10], and the minimum number of samples per split from [2,5,\underline{10}]. For the LSTM model, we tune the learning rate from [1e-4,\underline{5e-4},1e-5], batch size [\underline{16},32], number of LSTM layers from [1,\underline{2}], dropout from [0, \underline{0.1}], and fix the number of hidden dimensions to 128. For the BERT model, we tune the learning rate from [3e-5, \underline{3e-6}, 3e-7] and fix the batch size to 8.


\begin{table*}[]
    \centering
    \resizebox{0.99\linewidth}{!}
     {
\begin{tabular}{l|ccc|ccc||ccc|ccc}
\toprule
    &  \multicolumn{6}{c||}{Train: Snopes} & \multicolumn{6}{c}{Train: PolitiFact} \\
    & \multicolumn{3}{c|}{Within dataset} & \multicolumn{3}{c||}{Out-of dataset} & \multicolumn{3}{c|}{Within dataset} & \multicolumn{3}{c}{Out-of dataset} \\
    & \multicolumn{3}{c|}{Eval: Snopes} & \multicolumn{3}{c||}{Eval: PolitiFact} & \multicolumn{3}{c|}{Eval: PolitiFact} & \multicolumn{3}{c}{Eval: Snopes} \\
    RF & $\textrm{acc}_{\textrm{false}}$ & $\textrm{acc}_{\textrm{mix}}$ & $\textrm{acc}_{\textrm{true}}$ & $\textrm{acc}_{\textrm{false}}$ & $\textrm{acc}_{\textrm{mix}}$ & $\textrm{acc}_{\textrm{true}}$ & $\textrm{acc}_{\textrm{false}}$ & $\textrm{acc}_{\textrm{mix}}$ & $\textrm{acc}_{\textrm{true}}$ & $\textrm{acc}_{\textrm{false}}$ & $\textrm{acc}_{\textrm{mix}}$ & $\textrm{acc}_{\textrm{true}}$  \\ \midrule
Claim & 0.710 & 0.144 & 0.255 & \underline{0.853} & \underline{0.016} & \underline{0.209}  & 0.623 & 0.216 & \underline{0.513} & 0.790 & \underline{0.092} & \underline{0.255} \\
Evidence & 0.705 & \underline{0.152} & 0.441 & 0.829 & 0.006 & 0.117  & \underline{0.654} & 0.248 & 0.510 & \textbf{\underline{0.891}} & 0.039 & 0.192  \\
Claim+Evidence & \textbf{\underline{0.760}} & 0.136 & \underline{0.453} & 0.829 & 0.000 & 0.117  & 0.634 & \underline{0.292} & 0.512 & 0.871 & 0.039 & 0.199  \\
\midrule
\multicolumn{12}{l}{LSTM} \\
\midrule
Claim & 0.674 & 0.232 & \underline{0.280} & 0.875 & \underline{0.047} & \underline{0.137}  & 0.566 & 0.212 & \underline{0.504} & \underline{0.833} & 0.026 & 0.234   \\
Evidence & 0.721 & \underline{0.272} & 0.267 & \textbf{\underline{0.890}} & 0.020 & 0.115  & 0.643 & \underline{0.253} & 0.485 & 0.768 & \underline{0.184} & 0.322   \\
Claim+Evidence & \underline{0.757} & 0.248 & 0.168 & 0.879 & 0.008 & 0.107  & \textbf{\underline{0.671}} & 0.210 & 0.460 & 0.704 & 0.171 & \textbf{\underline{0.378}}   \\
\midrule
\multicolumn{12}{l}{BERT} \\
\midrule
Claim & 0.746 & 0.256 & 0.379 & 0.854 & \textbf{\underline{0.094}} & 0.045  & 0.604 & 0.292 & 0.475 & 0.765 & 0.171 & 0.287  \\
Evidence & 0.648 & \textbf{\underline{0.376}} & \textbf{\underline{0.559}} & 0.702 & 0.049 & \textbf{\underline{0.337}}  & 0.649 & \textbf{\underline{0.326}} & 0.496 & \underline{0.804} & \textbf{\underline{0.197}} & 0.339  \\
Claim+Evidence & \underline{0.747} & 0.264 & 0.379 & \underline{0.882} & 0.067 & 0.042  & \underline{0.667} & 0.175 & \textbf{\underline{0.558}} & 0.790 & 0.092 & \underline{0.367}  \\
\bottomrule
\end{tabular}}
    \caption{Accuracy scores computed on the false labels, mixture or half-true label, and true labels. All labels within a group (e.g., any false label such as false or mostly false) are considered to be the same and as such this reduces the problem to a three class classification problem.}
    \label{tab:results-per-veracity}
\end{table*}

\subsection{Results}
The results can be seen in Table \ref{tab:results}. Overall, we see that the BERT model trained only on Evidence obtains the best results in 4/8 columns, and, notably, in 3/4 cases the BERT model with Evidence obtains the best macro F1 score on within and out-of dataset prediction.
Random forest using term-frequency as input obtains the best out-of dataset micro F1 for both datasets (using either only Claim or only Evidence). Across all methods, the combination of Claim+Evidence only marginally obtains the best results a single time (for Snopes micro F1). For further details, in Table \ref{tab:results-per-veracity} we compute the accuracy scores for all the false labels, mixture or half-true label, and true labels.

Surprisingly, a BERT model using only the Evidence is capable of predicting the veracity of the claim used for obtaining the evidence. This shows that a strong signal must exist in the evidence itself, and the evidence found by the search engine appears to be implicitly affected by the veracity of the claim used as the query in some way\footnote{Note that the claim origin website is always removed from the evidence.}. The improvements reported in the literature by combining claim and evidence, are therefore not evident of the model learning to reason over the evidence with regards to the claim, but instead exploiting a signal inherent in the evidence itself. This highlights that the current approach for evidence gathering is problematic, as the strong signal makes it possible (and most often beneficial) for the model to entirely ignore the claim. This makes the model entirely reliant on the process behind how the evidence is generated, which is outside the scope of the model, and thereby undesirable, as any change in the search system may change the model performance significantly. 
It may also be problematic on a more fundamental level, e.g., to predict the veracity of the following two claims: ''the earth is round'' and ''the earth is flat'', the evidence could be the same, but a model entirely dependent on the evidence, and not the claim, would be incapable of predicting both claims correctly.

\subsection{Removal of evidence} We observed a strong predictive signal in the evidence alone and now consider the performance impact when gradually removing evidence snippets. The evidence is removed consecutively either from the top down or bottom up (i.e., removing the most relevant snippets first and vice versa), until no evidence is used. Figure \ref{fig:plots} shows the macro F1 as a function of removed evidence when using the Evidence or Claim+Evidence models. We observe a distinct difference between the random forest and LSTM model compared to BERT: for random forest and LSTM, the Claim+Evidence models on both datasets drop rapidly in performance when the evidence is removed, while the BERT model only experiences a very small drop. This shows that when the Claim+Evidence is used in the BERT model, the influence of the evidence is minimal, while the evidence is vital for the Claim+Evidence RF and LSTM models. For all models, we observe that when evidence is removed from the top down, the performance drop is larger than when evidence is removed from the bottom up. Thus, the ranking of the evidence as provided by the search engine is related to its usefulness as evidence for fact checking.

\section{Conclusion}
We investigate if fact checking models for fake news detection are learning to process claim and evidence jointly in a way resembling reasoning. Across models of varying complexity and evaluated on multiple datasets, we find that the best performance can most often be obtained using only the evidence. This highlights that models using both claim and evidence are inherently not learning to reason, and points to a potential problem in how evidence is currently obtained in existing approaches for automatic fake news detection.


\bibliographystyle{acl_natbib}
\bibliography{anthology,acl2021}


\end{document}